%% file: main.tex
\documentclass[10pt,twocolumn,letterpaper]{article}
\newcommand{\ours}{SGVLM}
\newcommand{\momaqa}{MOMA-QA}

\usepackage{cvpr}              
\usepackage[utf8]{inputenc}
\usepackage{pifont}
\usepackage{tabularx,stackengine}
\newcolumntype{L}[1]{>{\minwd l{#1}}l<{\endminwd}}
\newcolumntype{C}[1]{>{\minwd c{#1}}c<{\endminwd}}
\newcolumntype{R}[1]{>{\minwd r{#1}}r<{\endminwd}}
\def\minwd#1#2#3\endminwd{\stackengine{0pt}{#3}{\rule{#2}{0pt}}{O}{#1}{F}{F}{L}}
\newcolumntype{Q}{@{}c@{}}
\usepackage{booktabs}
\usepackage{multirow}
\usepackage{makecell}
\usepackage{amssymb}
\usepackage{xr-hyper}

\input{preamble}

\definecolor{cvprblue}{rgb}{0.21,0.49,0.74}
\usepackage[pagebackref,breaklinks,colorlinks,citecolor=cvprblue]{hyperref}

\def\paperID{10943} 
\def\confName{CVPR}
\def\confYear{2024}

\title{Towards Fine-Grained Video Question Answering}

\author{Wei Dai, Alan Luo, Zane Durante, Debadutta Dash, Arnold Milstein, Kevin Schulman, \\ Ehsan Adeli, Li Fei-Fei \\
Stanford University\\
{\tt\small \{dvd.ai, alanzluo, durante, ddash, amilstein, Kevin.schulman, eadeli\}@stanford.edu,} \\
{\tt\small feifeili@cs.stanford.edu}
}

\begin{document}
\maketitle
\input{sec/0_abstract}    
\input{sec/1_intro}
\input{sec/2_related_works}
\input{sec/3_moma_qa_dataset}
\input{sec/4_method}
\input{sec/5_experiments}
\input{sec/6_discussion}
\input{sec/7_conclusion}
{
    \small
    \bibliographystyle{ieeenat_fullname}
    \bibliography{main}
}

\input{sec/X_suppl}

\end{document}

%% file: preamble.tex
\usepackage[dvipsnames]{xcolor}


%% file: sec/0_abstract.tex
\begin{abstract}
In the rapidly evolving domain of video understanding, Video Question Answering (VideoQA) remains a focal point. However, existing datasets exhibit gaps in temporal and spatial granularity, which consequently limits the capabilities of existing VideoQA methods. This paper introduces the Multi-Object Multi-Actor Question Answering (MOMA-QA) dataset, which is designed to address these shortcomings by emphasizing temporal localization, spatial relationship reasoning, and entity-centric queries. With ground truth scene graphs and temporal interval annotations, MOMA-QA is ideal for developing models for fine-grained video understanding. Furthermore, we present a novel video-language model, SGVLM, which incorporates a scene graph predictor, an efficient frame retriever, and a pre-trained large language model for temporal localization and fine-grained relationship understanding. Evaluations on MOMA-QA and other public datasets demonstrate the superior performance of our model, setting new benchmarks for VideoQA.
\end{abstract}

%% file: sec/1_intro.tex
\vspace{-1em}
\section{Introduction}


In the current era of abundant digital video content, video understanding has become a key focus in computer vision research, with significant implications in various fields such as entertainment \cite{tapaswi2016movieqa, huang2020movienet, rao2020local}, healthcare \cite{singh2020automatic, rajavel2022iot, khan2020smsh, shorfuzzaman2021towards}, and surveillance \cite{sreenu2019intelligent, ullah2021cnn}. Among the numerous aspects of video comprehension, Video Question Answering (VideoQA) has garnered a significant amount of attention, since it requires models to answer questions regarding a specific video segment, which necessitates a thorough grasp of the scene, relationships, and temporal changes depicted in the video \cite{zhong-etal-2022-video,jang2017tgif,xiao2021next}. 

Grounding in video understanding—specifically, temporal and spatial grounding—plays a pivotal role in bridging the gap between low-level video features and high-level semantic interpretations \cite{wang-etal-2023-vstar}. Temporal grounding ensures that events or actions within videos are associated with specific time intervals \cite{soldan2021vlg}, while spatial grounding offers localized regions within video frames that correspond to certain entities or objects \cite{tan2021look}. A dataset incorporating both temporal and spatial dimensions can offer rich contextual cues and pave the way for more detailed and accurate video comprehension tasks.

\begin{figure}
    \centering
    \includegraphics[width=\linewidth]{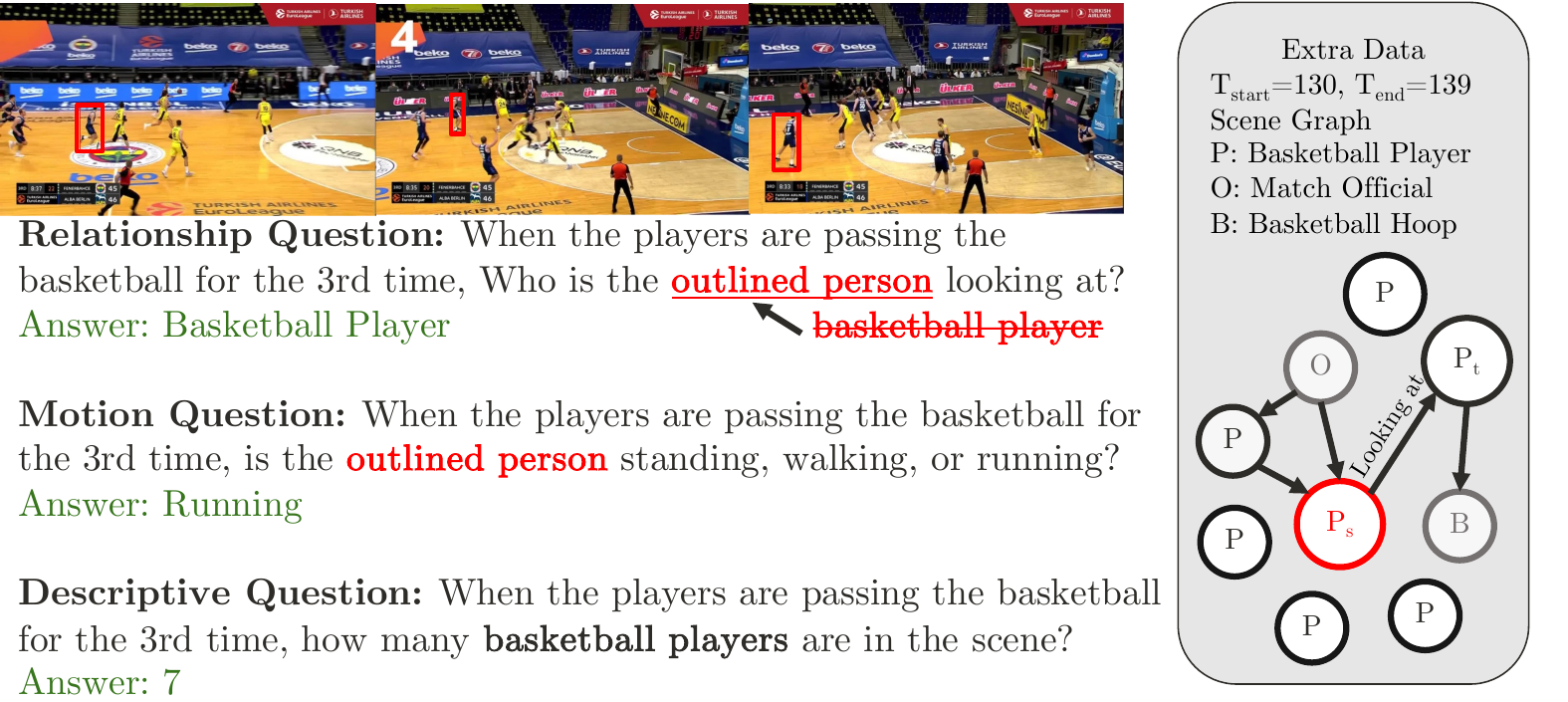}
    \caption{\textbf{Visualizations of Sample Questions from \momaqa.} We illustrate the three distinct types of questions in our dataset, each representing a different category for video question answering. All questions in our dataset are generated from a human-annotated spatio-temporal scene graph (shown on the right).  The node of interest for the relationship and motion questions is colored \textcolor{red}{red} in the scene graph and outlined in the video.}
    \label{fig:dataset_vis}
    \vspace{-1em}
\end{figure}

While many datasets exist in the realm of video understanding, a deeper dive into their contents uncovers notable gaps in their representations of spatial relationships. For instance, in the ActivityNet-QA dataset \cite{yu2019activitynet}, only 10\% of its questions revolve around questions with spatial dimensions. This restricts the range and depth of inquiries a model can proficiently address. Another concern is the absence of fine-grained spatial relationship annotations. While TVQA+ \cite{lei2019tvqa+} offers object-level details via bounding boxes, it fails to provide relationships between these objects. STAR \cite{wu2021star_situated_reasoning} provides relationship annotations for its videos,  but the automated nature of these annotations significantly restricts their precision and applicability.

\input{tables/qa_gen}

Temporal grounding, too, has its share of challenges in existing VideoQA datasets. Foundational datasets such as MSRVTT-QA \cite{xu2017video} often neglect the importance of temporal localization. Approximately 33.4\% of its questions can be distilled to a generic format: ``What is [someone] doing'' Such questions steer models predominantly toward video classification objectives, bypassing the need to anchor responses to specific moments or sequences within a video. Recent datasets like TVQA+ and TGIF-QA \cite{lei2019tvqa+, jang2017tgif} have shifted focus towards temporal reasoning within videos. However, they lack ground truth annotations for temporal localizations, thus there is no definitive means to ascertain whether a model has accurately localized the correct frames. Lastly, understanding and identifying the actions of individuals in crowded settings is challenging, and few datasets tackle this \cite{luo2022moma}. Addressing entity-specific queries in group situations is crucial for advanced video comprehension.



Given the gaps observed in current datasets, we introduce the Multi-Object Multi-Actor Question Answering (\momaqa) dataset. Stemming from the foundation of the \textit{Multi-Object Multi-Actor (MOMA)} \cite{luo2022moma} dataset, MOMA-QA brings unique attributes designed to challenge and improve the current generation of video question-answering models. Firstly, as shown in Figure \ref{fig:dataset_vis}, every question within MOMA-QA requires temporal localization and is accompanied by ground truth temporal interval annotations to provide a means to validate models' temporal localization abilities. Secondly, 71.6\% of the questions in the dataset require spatial relationship understanding, which MOMA-QA intensively assesses models on interpreting spatial connections among video entities. Each frame features ground truth scene graph annotations, laying a foundation for the evolution of more sophisticated spatially-aware models. Lastly, understanding the challenge of discerning specific individuals in crowded settings, we visually demarcate specific actors via frame-level bounding boxes on a subset of questions, thereby testing the model's proficiency in entity-specific reasoning.

As current datasets lack fine-grained annotations, existing VideoQA models struggle with nuanced understanding due to their linear approach of directly processing video frames and questions to produce answers. This limits their interpretability, a gap that becomes more apparent with the rise of visual language models in this domain. To address this issue, we introduce \textbf{\ours}, a video-language model with enhanced retrieval and relationship understanding abilities. Our model encapsulates three main features. First, the vision encoder has been restructured and augmented with a Motif-based scene graph generator \cite{zellers2018neural}. The scene graph generator provides robust grounding for the spatial relationships depicted in videos and also provides an interpretable understanding of the model's decision-making pathway and elucidating how it arrives at its final predictions. Second, we devise an efficient frame retriever that identifies frames relevant to posed questions by leveraging both video and scene graph features, providing greater accuracy, especially for tasks on discerning relationships. Lastly, \ours~ harnesses the power of pre-trained large language models, empowering it to tackle intricate reasoning tasks.

In summary, our work has the following contributions: (1)
%
    We present the \textbf{MOMA-QA} dataset, a VideoQA dataset that emphasizes temporal localization, relationship reasoning through a vast array of questions, and frame-level entity-specific annotations to enhance video question-answering models. Each question is equipped with ground truth relationship and temporal annotation to facilitate the development of fine-grained VideoQA models. 
    (2)
    We introduce \textbf{\ours}, a video-language model that features a restructured vision encoder with a Motif-based scene graph generator for spatial relationship grounding, an efficient frame retriever for selecting relevant frames, and the integration of pre-trained large language models for advanced reasoning capabilities.

%% file: tables/qa_gen.tex
\begin{table*}
    \centering
    \small
    \resizebox{\linewidth}{!}{%
    \begin{tabular}{@{}lp{4cm}l p{7.3cm} p{2.3cm}@{}}
        \toprule
         \textbf{Category} &  \textbf{Question Format} & \textbf{Answer Format} &  \textbf{Example Question} & \textbf{Example Answer}   \\
         \midrule
         Relationship & When $[C_i]$, what is $[V_{s}]$ $[E_i]$? & $[V_t]$ & When the players are passing the basketball for the 3rd time, Who is the outlined person looking at? & Basketball Player \\
         \midrule
         Motion & When $[C_i]$, is $[V_t]$ standing, walking or running? & [Att($V_t$)] & When the players are passing the basketball for the 3rd time, is the outlined person standing, walking, or running? & Running \\
         \midrule
         Description & When $[C_i]$, how many \text{[Identity]} are in the scene? & [Id. Count] & When the players are passing the basketball for the 3rd time, how many basketball players are in the scene? & 7 \\
         \bottomrule
    \end{tabular}
    }
    \caption{\textbf{General Structure of Generated Questions.} $C_i$ denotes the description of a particular sub-activity. Additionally, $V_s$ and $V_t$ denotes the name of a source and a target node from the sub-activity. $E_i$ represents the description of a relationship connecting $V_s$ and $V_t$.}
    \label{tab:qa_gen}
    \vspace{-1em}
\end{table*}

%% file: sec/2_related_works.tex
\section{Related Works}

\input{tables/dataset}

\noindent\textbf{Video Question Answering Datasets.} The quality of machine learning models is heavily influenced by the quality of their datasets. Foundational datasets like MovieQA \cite{movieqa}, MSRVTT-QA, and MSVD-QA \cite{xu2017video} have significantly advanced video question answering research \cite{PengWG0Z22, LeiLZGBB021, justask, hga}. However, these datasets mainly include short clips with simple questions, limiting the development of models' in-depth video understanding. TGIF-QA \cite{jang2017tgif} introduced a significant change by testing spatial-temporal reasoning in a large dataset of animated GIFs, leading to improvements in models' temporal reasoning abilities \cite{comem, HME}. Despite this, there remains a gap in spatial reasoning and the ability to handle crowded scenes with similar-looking actors. 

\noindent\textbf{Grounded VideoQA Models.} Grounding in VideoQA tasks usually consists of two parts: spatial and temporal. With the advent of graph neural networks (GNN) \cite{gcn}, many works \cite{LGCN, HAIR, PGAT, MASN} have integrated GNNs within their VideoQA framework for spatial grounding. Temporal grounding involves identifying salient frames related to the input question \cite{buch2022revisiting}. This technique gained increasing attention as LLM-based models became popular \cite{alayrac2022flamingo, touvron2023llama,flant5}. While LLM-based models bolster advanced reasoning abilities, their input lengths are strictly capped, making advance frame selection necessary \cite{yu2023self}. This paper presents the first LLM-based VideoQA model that utilizes both temporal and spatial grounding features.

%% file: tables/dataset.tex
\begin{table*}[htbp!]
\small
\centering
\resizebox{\textwidth}{!}{%
    \begin{tabular}{@{}lcccccccc@{}}
        \toprule
        \textbf{Dataset} & \textbf{Video Source} & \textbf{\#Videos} & \textbf{\makecell{\#QA \\ Pairs}} & \textbf{\makecell{Average \\ Length (s)}} & \textbf{\makecell{Open \\ Ended}} & \textbf{\makecell{Temporal \\ Localization}} & \textbf{\makecell{Bounding Box \\ Augmentation}} & \textbf{\makecell{Scene Graph \\ Annotation}} \\
        \midrule
         MSVD-QA~\cite{xu2017video}& MSVD  & $1,970$ & $50,505$ & $10$ & \checkmark & \ding{55} & \ding{55} & \ding{55} \\
         MSRVTT-QA~\cite{xu2017video}& MSRVTT  & $10,000$ & $243,690$ & $15$ & \checkmark & \ding{55} & \ding{55} & \ding{55} \\
         TGIF-QA~\cite{jang2017tgif}& TGIF  & $71,741$ & $165,165$ & $3$ & \checkmark & \checkmark & \ding{55} & \ding{55} \\
         TVQA~\cite{lei2018tvqa} & TV Show  & $21,793$ & $152,545$ & $76$ & \ding{55} & \checkmark & \ding{55} & \ding{55} \\
         ActivityNet-QA~\cite{yu2019activitynet} & ActivityNet  & $5,800$ & $58,000$ & $180$ & \checkmark & \checkmark & \ding{55} & \ding{55} \\
         Social-IQ~\cite{zadeh2019social}&YouTube  & $1,250$ & $7,500$ & $60$ & \ding{55} & \ding{55} & \ding{55} & \ding{55} \\
         EgoSchema \cite{mangalam2023egoschema} & Ego4D & $5,063$ & $5,063$ & $180$ & \ding{55} & \ding{55} & \ding{55} & \ding{55} \\
         NExT-QA~\cite{xiao2021next} & YFCC-100M & $5,440$ & $52,044$ & $44$ & \checkmark & \checkmark & \ding{55} & \ding{55} \\
         STAR~\cite{wu2021star_situated_reasoning} & Charades & $23,013$ & $60,206$ & $11$ & \ding{55} & \checkmark & \ding{55} & $\bigcirc^*$ \\
         TVQA+~\cite{lei2019tvqa+} & TV Show  & $4,198$ & $29,383$ & $61$ & \checkmark & \checkmark & \checkmark & \ding{55} \\
         \midrule
         MOMA-QA (Raw only) & MOMA & $1,412$ & $83,223$ & $376$ & \checkmark & \checkmark & \ding{55} & \checkmark \\
         MOMA-QA (Aug.) & MOMA & $27,586$ & $300,791$ & $144$ & \checkmark & \checkmark & \checkmark & \checkmark \\
        \bottomrule
    \end{tabular}
    }
    \caption{\textbf{Dataset Comparisons.} Our proposed dataset sets a new benchmark for open-ended, long VideoQA by providing extensive human annotations and a large number of QA pairs. Raw only: Includes raw videos only. Aug.: Includes both raw videos and box-augmented videos. * The graph annotations provided by the STAR dataset are automatically generated, while \momaqa's are human annotated.
    \label{tab:dataset_comp}
    }
\end{table*}

%% file: sec/3_moma_qa_dataset.tex
\section{MOMA-QA Dataset}

In this section, we introduce the MOMA-QA dataset through three perspectives: source annotations, questions, and its feature of bounding box augmentations. We perform the same video-wise train/validation/test split as in the MOMA dataset. We then show the statistics of \momaqa~and compare it with the current VideoQA datasets. 

\subsection{Annotations}

The MOMA dataset contains human annotated \textit{activity graphs} at the frame level. Specifically, each frame $i$ is annotated with a graph $G_i = (V_i, E_i)$, where $V_i$ contains a set of entities, along with their bounding boxes and attributes in the scene. $E_i$ contains the relationships between the entities. In addition, consecutive frames are grouped into sub-activities. Sub-activity $j$ has label $(T_{start, j}, T_{end, j}, C_j)$, where $T_{start, j}, T_{end, j}$ denotes the start and end of a particular activity, and $C_j$ contains the description about the sub-activity. Such fine-grained human annotations make it ideal for question generation on relationships while also providing extra information for model grounding.

\subsection{Bounding Box Augmentations}

Analyzing crowded scenes poses challenges like question ambiguity. Consider the inquiry: ``What is the basketball player looking at?'' Posed within the context of a match involving ten participants, it becomes unclear to which player the question is directed.  Nonetheless, entity-centric reasoning within such crowded scenes is critical across various domains. For instance, in a sports event, the analysis of a particular player's performance gathers considerable interest. To address this issue, we propose \textit{bounding box augmentations}. This technique generates edited videos highlighting the focused entity using ground truth bounding box annotations from the MOMA dataset, as shown in Figure~\ref{fig:dataset_vis}. This method effectively resolves the ambiguity, thereby facilitating entity-centric reasoning within dense scenes. Furthermore, we substitute specific entity designations (like ``basketball player'') with the more generic term ``\textit{outlined person}.'' This alteration serves to minimize the hints the question may provide regarding the answer, thus preventing the model from inferring the answer based solely on the phrasing of the question. For fair comparison to existing works, we do not supply any bounding box coordinates during QA.

\subsection{Questions}

In the \momaqa~dataset, we offer three categories of questions: \textit{relationship}, \textit{motion}, and \textit{description}. The standard template used to generate these questions is outlined in Table~\ref{tab:qa_gen}. After generation, each question undergoes a manual verification process to confirm its clarity and remove any ambiguity. Adjustments are made to the phrasing of questions to enhance their naturalness.

Figure~\ref{fig:dataset_vis} presents exemplar questions from the \momaqa~dataset. Specifically, every question is accompanied by precise interval and scene graph ground truth annotations. We hope the inclusion of this information could drive the development of more intricate multimodal models. 

\subsection{Dataset Statistics}

Table~\ref{tab:dataset_comp} presents a comparative analysis of the \momaqa~dataset against other popular VideoQA datasets. The \momaqa~dataset stands out with its extensive collection of 300,791 questions derived from 147 hours of original footage and an additional 956 hours of bounding-box augmented videos, making it one of the most comprehensive VideoQA datasets currently available. Furthermore, the average duration of 144 seconds per video makes this dataset well-suited for evaluating long-form temporal localization in models. In addition, \momaqa~is one of the first datasets to provide human annotated temporal interval and scene graph data within the VideoQA domain.

\begin{figure}
    \centering
    \includegraphics[width=\linewidth]{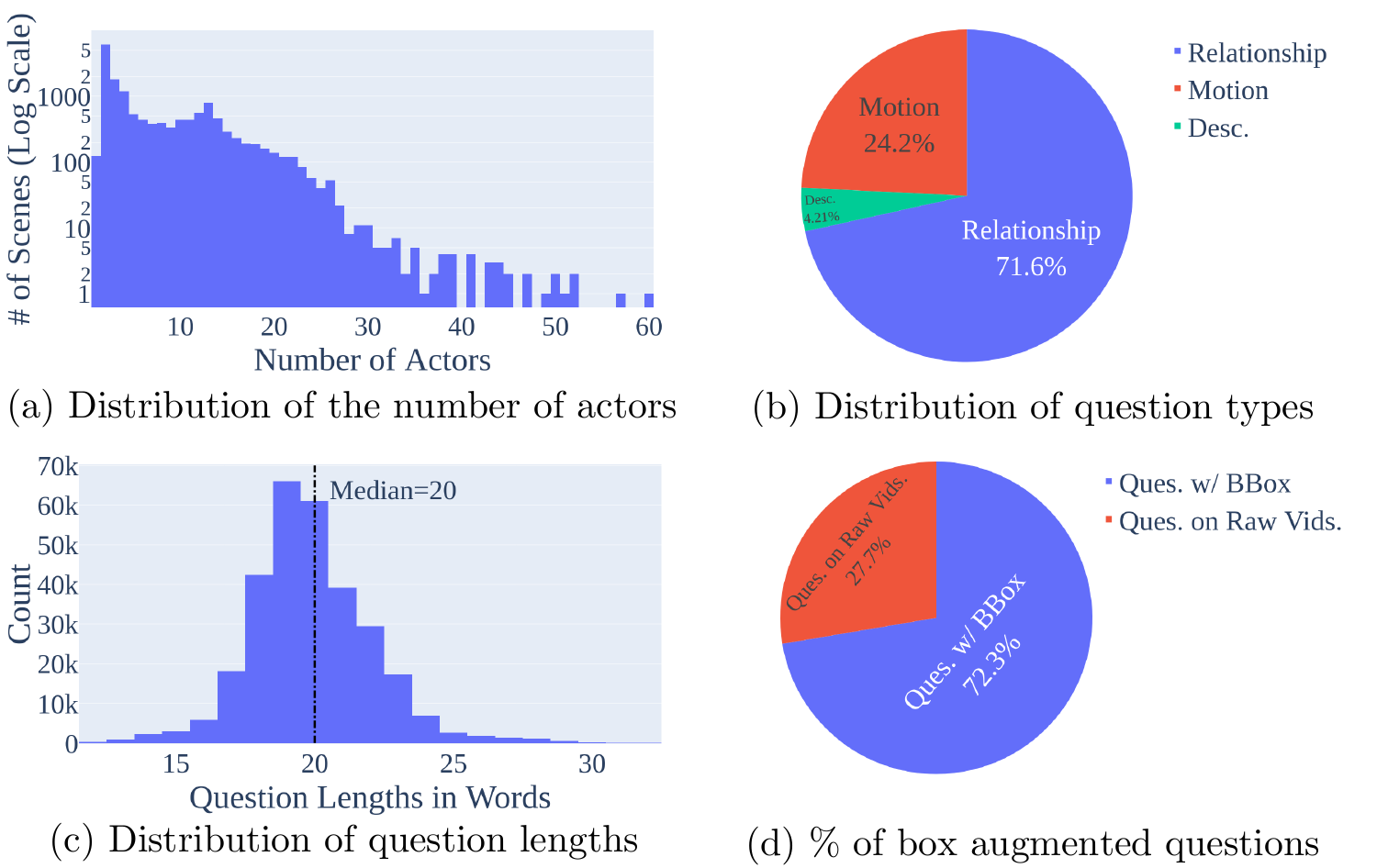}
    \caption{\textbf{Statistics of \momaqa.} (a) The distribution of the number of actors. (b) The percentage of each question type in \momaqa. (c) The distribution of question lengths in \momaqa~in words. (d) The percentage of box-augmented questions.}
    \label{fig:moma_stats}
    \vspace{-1em}
\end{figure}

Statistics from Figure~\ref{fig:moma_stats} reveal that 71.6\% of the questions in the dataset are centered on relationships, 24.2\% pertain to motion, and the remaining 4.21\% are descriptive. This distribution underscores the dataset’s focus on relational understanding. Additionally, the dataset exhibits a balanced distribution of question lengths, with a median count of 20 words per question. It contains 4,045 scenes that contain over 10 actors each, and 72.3\% of the questions have been enhanced with bounding box annotations, emphasizing the dataset's dedication to entity-specific queries.

%% file: sec/4_method.tex
\section{Method}

\begin{figure*}
    \centering
    \includegraphics[width=0.9\textwidth]{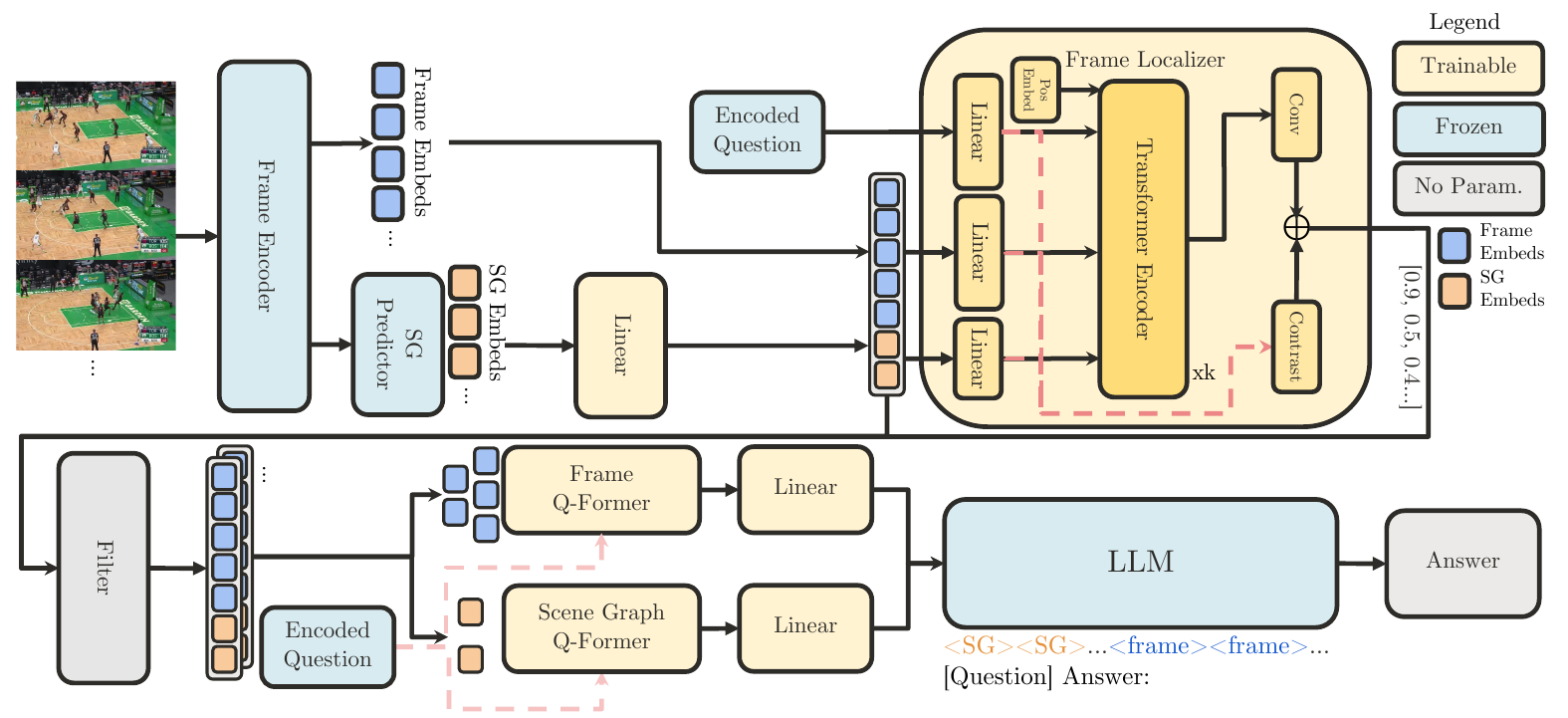}
    \caption{\textbf{Model Architecture of \ours.} The model employs a frame encoder to extract frame embeddings from the input video, which are subsequently used by a Scene Graph (SG) Predictor to generate scene graph embeddings. These embeddings are then concatenated with the frame features. The combination, along with question embeddings, is processed by a transformer encoder in the Frame Localizer to produce similarity scores for identifying relevant frames. Key frame features are then processed by Frame Q-Former and SG Q-Former to align with the language query and scene graph features. An LLM finally generates answers using a structured representation of scene graph and frame data, merged with the natural language question.}
    \label{fig:model_arch}
    \vspace{-1em}
\end{figure*}


As shown in Figure \ref{fig:model_arch}, \ours~fuses video frame features with language for advanced video understanding. Our model has five main components: frame encoder, scene graph predictor, frame localizer, Q-Former, and LLM. An illustration of each component is detailed below. 

\noindent\textbf{Frame Encoder.} We utilize EVA-02 \cite{EVA02}, a ViT based image encoder with 304M parameters, to generate patch image features $X \in \mathbb{R}^{L_{patch} \times d_v}$, object bounding box predictions $B=\{b_1, \dots, b_n\}$, object class predictions $O=\{o_1, \dots, o_n\}$, and image features used for scene graph generation. In particular, besides the coordinates of each box proposal prediction, the bounding box prediction $B$ also contains a feature vector $\mathbf{f}_i$ and label probability $\mathbf{p}_i$ for each proposal $b_i \in B$. We utilize the pre-trained weight from the original work and fine-tune it on Visual Genome \cite{krishna2017visual} and \momaqa. 

\noindent\textbf{Scene Graph Predictor.} We design our scene graph predictor based on the Neural Motifs structure \cite{zellers2018neural} to pre-train our frame encoder on Visual Genome \cite{krishna2017visual}. We use biLSTM layers to encode object and edge contexts, which are then used to build relationship features. The features with top k probabilities $\mathbf{S} = \{\mathbf{s}_i, \dots, \mathbf{s}_k\}$ are extracted and used in subsequent steps. A detailed explanation of the process is included in Suppl.~\ref{sec:sup_sg}.

\begin{figure}
    \centering
    \includegraphics[width=0.9\linewidth]{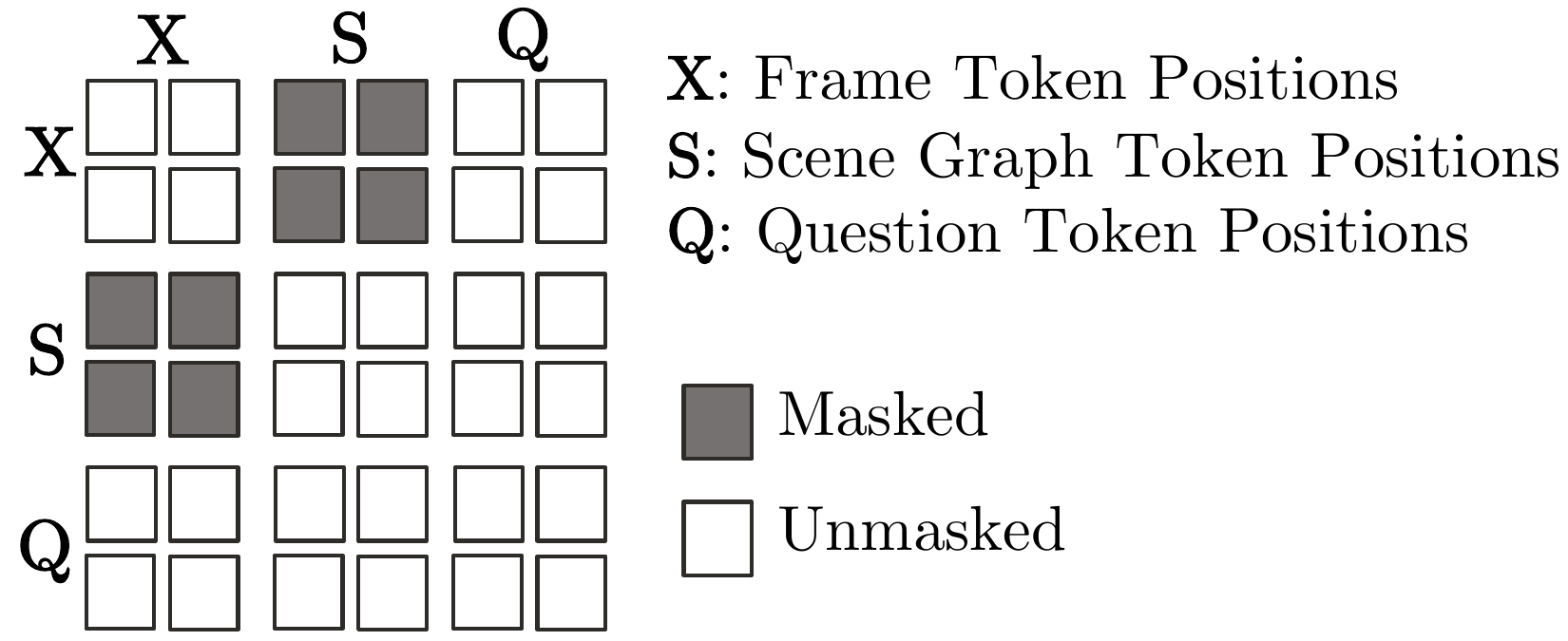}
    \caption{\textbf{Self-Attention Mask of the Transformer Encoder in Frame Localizer.} To separate frame and scene graph tokens, we mask out portions of the input with $-\infty$. }
    \label{fig:att_mask}
    \vspace{-1em}
\end{figure}

\noindent\textbf{Frame Localizer.} The frame localizer, based on the UniVTG \cite{lin2023univtg} structure, employs a hybrid alignment and contrastive approach, leveraging frame and scene graph embeddings. During training, frames are labeled with binary $f_i$, where $f_i=1$ signifies a foreground clip, and a saliency score $s_i \in [-1, 1]$, indicating relevance to the target question. The input question is converted into query tokens $\mathbf{Q} \in \mathbb{R}^{n \times d_t}$. For each frame, the frame embedding $\mathbf{X}$ and scene graph embedding $\mathbf{S}$ undergo a separate linear layer:
\begin{align*}
    \mathbf{x}_{i} = \frac{1}{|\mathbf{X_i}|}\sum_{j=1}^{|\mathbf{X_i}|} \mathbf{X_i}\mathbf{W}_{xs}, \quad \mathbf{s}_{i} = \frac{1}{|\mathbf{S_i}|}\sum_{j=1}^{|\mathbf{S_i}|} \mathbf{S_i}\mathbf{W}_{ss},
\end{align*}
where $\mathbf{W}_{xs}, \mathbf{W}_{ss}$ are learnable matrices. The squashed frame embedding and scene graph embeddings are then separately concatenated to form video frame embedding $\mathbf{X}_v = \{\mathbf{x}_1, \dots, \mathbf{x}_n\}$ and video scene graph embedding $\mathbf{S}_v = \{\mathbf{s}_1, \dots, \mathbf{s}_n\}$ for video of length $n$. In the alignment route, each modality is appended with the positional embedding and type embeddings: $\mathbf{X}_v' = \mathbf{X}_v + \mathbf{E}_{X}^{pos} + \mathbf{E}_{X}^{type} ;
    \mathbf{Q}_v' = \mathbf{Q}_v + \mathbf{E}_{Q}^{pos} + \mathbf{E}_{Q}^{type} ;
    \mathbf{S}_v' = \mathbf{S}_v + \mathbf{E}_{S}^{pos} + \mathbf{E}_{S}^{type}$. Then, the embeddings concatenated into $\mathbf{Z}_0 = [\mathbf{X}_v';\mathbf{S}_v';\mathbf{Q}_v']$. The concatenated representation $\mathbf{Z}_0$ is fed into a stack of $k$ transformer encoders, where each encoder is composed of a multi-head self-attention (MHSA) and a linear layer. At layer $i$ with an MHSA with $m$ heads, we have
\begin{align*}
    \mathbf{k}_{i,m} &= \text{softmax}\left(\frac{\mathbf{W}_Q^{i,m} \mathbf{Z}_{i-1} (\mathbf{W}_K^{i,m} \mathbf{Z}_{i-1})^T}{\sqrt{d_k^i}} + \mathbf{M} \right), \\
    \mathbf{h}_{i,m} &= \mathbf{k}_{i,m} \mathbf{W}_V^{i,m} \mathbf{Z}_{i-1}, \\
    \mathbf{Z}_i &= (||_{m=1}^{M} \mathbf{h}_{i,m} ) \mathbf{W}^i_{O},
\end{align*}
where $\mathbf{W}^i_{O}, \mathbf{W}_Q^{i,m}, \mathbf{W}_K^{i,m}, \mathbf{W}_V^{i,m}$ are learnable parameters, and $\mathbf{M}$ is the attention mask as configured in Figure \ref{fig:att_mask}. 
Our architecture implements a specialized attention mask that restricts frame and scene graph tokens to interact exclusively with question tokens. This design choice is grounded in the empirical finding that scene graph and frame tokens exhibit inherent correlation. Without this masking, the localizer shows a propensity to assign high attention scores to the interplay between frame and scene graph tokens, often at the expense of question token relevance. By enforcing this attention mask, we ensure that the focus remains on integrating the question context effectively, as demonstrated by Suppl. \ref{sec:att_masks}. In the end, we remove the question token part of $\mathbf{Z}_k$ and leave only the frame and scene graph tokens to obtain $\mathbf{Z}_{k}'$. The final score for the alignment route is then obtained by
\begin{equation}
    \mathbf{\hat{f}} = \sigma(\text{Conv}(\mathbf{Z}'_k)),
\end{equation}
where $\sigma$ is a sigmoid activation, and Conv is a set of convolutional layers that outputs $\mathbf{\hat{f}} \in \mathbb{R}^{n} = \{\hat{f}_{1}, \dots, \hat{f}_{n}\}$, where each value predicts whether the frame belongs to a foreground clip. The alignment route is then supervised by the cross entropy loss between the predicted label $\mathbf{\hat{f}}_{a}$ and the ground truth label $f_{a}$:
\begin{equation}
    \mathcal{L}_{a} = \sum_{i=1}^n - \left(f_{i} \text{log}~\hat{f}_{i} + (1-f_{i}) \text{log}(1 - \hat{f}_{i})\right),
\end{equation}
where $s_i$ is the ground truth relevance at frame $i$. 

In the contrastive learning route, a one-layer attention layer is first used to project the question embedding $\mathbf{Q}' = \text{softmax}(\mathbf{W}_c \mathbf{Q})\mathbf{Q}$ where $\mathbf{W}_c$ is a learnable parameter. Then, the saliency score $\hat{s}_c$ is obtained through the sum of the pair-wise similarity score between the frame embedding $\mathbf{S}_v = \{\mathbf{s}_1, \dots, \mathbf{s}_n\}$, scene graph embedding $\mathbf{X}_v = \{\mathbf{x}_1, \dots, \mathbf{x}_n\}$, and question embedding $\mathbf{Q}'$:
\begin{equation}
    \hat{s}_{c, i} = \frac{\mathbf{x}_i^T \mathbf{Q}'}{||\mathbf{x}_i||_2 ||\mathbf{Q}'||_2} + \frac{\mathbf{s}_i^T \mathbf{Q}'}{||\mathbf{s}_i||_2 ||\mathbf{Q}'||_2}.
\end{equation}
This score is supervised through two losses: intra-video and inter-video contrastive learning loss. For intra-video contrastive learning loss, we randomly sample a positive clip at index $p$ with $f_p=1$ and $s_p > 0$, and negative samples $N = \{j | 1 \leq j < p, s_j < s_p \}$. Given the saliency prediction $\hat{s}_j, \hat{s}_j$, the intra-video loss is calculated as
\begin{equation}
    \mathcal{L}_s^{\text{intra}} = -\log \frac{\exp(\hat{s}_p / \tau)}{\exp(\hat{s}_p / \tau) + \sum_{j \in N} \exp(\hat{s}_j / \tau)},
\end{equation}
where $\tau$ is a hyperparameter representing the temperature. The inter-video loss takes other videos $k \in N'$ within the batch as negative samples 
\begin{equation}
    \mathcal{L}_s^{\text{inter}} = -\log \frac{\exp(\hat{s}_p / \tau)}{\sum_{k \in B} \exp(\hat{s}_p^k / \tau)}.
\end{equation}
The overall training objective is the weighted combination:
\begin{equation}
    \mathcal{L} = \lambda_a \mathcal{L}_{a} + \lambda_{intra} \mathcal{L}_s^{\text{intra}} + \lambda_{inter} \mathcal{L}_s^{\text{inter}},
\end{equation}
where $\lambda_a, \lambda_{inter}, \lambda_{intra}$ are hyperparameters setting the weight for each loss. Finally, the relevance score $r_i$ for frame $i$ is the sum of both foreground prediction $\hat{f}_i$, and the saliency score $\hat{s}_i$:
\begin{equation}
    \hat{r}_i = w_f \hat{f}_i + w_s \hat{s}_i,
\end{equation}
where $w_f, w_s$ are two learnable scalars representing the weight of each score. The frames are ranked based on $\hat{r}_i$, and only the top k frames are input into the Q-Formers in the next stage. For datasets with ground truth interval annotation (like \momaqa), the localizer is directly tuned on the ground truth labels. For datasets without ground truth labels, pseudo labels $f'_i, s'_i$ are generated to fine-tune the localizer. Specifically, for each frame $i$ with answer prediction $y, \hat{y}$ and frame selection threshold $r_{\theta}$, 
\begin{equation}
    f'_i, s'_i =
    \begin{cases} 
        1, 1 & \text{if } (y = \hat{y} \wedge \hat{r}_i > r_{\theta}) \\ 
        & \text{~~~} \vee (y \neq\hat{y} \wedge \hat{r}_i < r_{\theta}) \\
        0, -1 & \text{Otherwise}
    \end{cases}.
\end{equation}
In other words, we encourage the localizer to make the same prediction if such prediction gives the correct answer while encouraging the model to make a different prediction when the selected frames fail to provide the correct answer. 

\noindent\textbf{Q-Formers and LLM.} We implement the Q-Formers as designed in BLIP-2 \cite{blip2}. In particular, since the scene graph and the frame embeddings have distinctively different embeddings, two Q-Formers are used, with one taking the frame embeddings and one taking the scene graph embeddings. A linear projection is then used to project the embedding into the LLM embedding space. Finally, the scene graph tokens, frame tokens and question tokens are concatenated and input into an LLM, and an LLM inference is performed to obtain the final answer. 

\input{tables/moma_zs}

\input{tables/moma_ft}

\input{tables/nextqa}

%% file: tables/moma_zs.tex
\begin{table*}[t]
\setlength{\tabcolsep}{0pt}
\setlength\extrarowheight{1pt}
\footnotesize
\begin{tabularx}{\linewidth}{l>{\centering\arraybackslash}X>{\centering\arraybackslash}XC{2mm}>{\centering\arraybackslash}X>{\centering\arraybackslash}XC{2mm}>{\centering\arraybackslash}X>{\centering\arraybackslash}XC{2mm}>{\centering\arraybackslash}X>{\centering\arraybackslash}X}
\toprule
\multirow{3}{*}{\vspace{0.1cm}\textbf{Model}} & \multicolumn{2}{c}{\textbf{Description}} & & \multicolumn{2}{c}{\textbf{Relationship}} && \multicolumn{2}{c}{\textbf{Action}} & & \multicolumn{2}{c}{\textbf{Total}} \\
\cmidrule{2-3} \cmidrule{5-6} \cmidrule{8-9} \cmidrule{11-12}
& \textbf{Accuracy} & \textbf{WUPS@0.9} & & \textbf{Accuracy} & \textbf{WUPS@0.9} & & \textbf{Accuracy} & \textbf{WUPS@0.9} & & \textbf{Accuracy} & \textbf{WUPS@0.9} \\
\midrule
InternVideo \cite{wang2022internvideo} & $0.00$ & $0.0000$ & & $0.07$ & $0.0018$ & & $0.00$ & $0.0000$ & & $0.05$ & $0.0013$ \\
mPLUG-2 \cite{mplug2} & $0.00$ & $0.5084$ & & $9.83$ & $0.2891$ & & $0.00$ & $0.0000$ & & $6.90$ & $0.2222$ \\
BLIP-2 \cite{blip2} & $\underline{55.19}$ & $0.5519$ & & $12.30$ & $0.1957$ & & $62.10$ & $0.6210$ & & $26.45$ & $0.3161$ \\
SeViLa \cite{yu2023self} & $53.22$ & $\textbf{0.6027}$ & & $\underline{12.99}$ & $\textbf{0.5045}$ & & $\underline{64.66}$ & $\underline{0.8977}$ & & $\underline{27.44}$ & $\textbf{0.6023}$ \\
\midrule
SGVLM & $\textbf{58.69}$ & $\underline{0.5869}$ & & $\textbf{13.03}$ & $\underline{0.4663}$ & & $\textbf{65.43}$ & $\textbf{0.9174}$ & & $\textbf{27.94}$ & $\underline{0.5828}$ \\
\bottomrule
\end{tabularx}
\caption{\textbf{Zero-Shot Performance Comparison of \ours~with Baselines on MOMA-QA Dataset.} Our method outperforms, or performs on-par with existing methods in the zero-shot setting. }
\label{tab:momaqa_zs}
\vspace{-1em}
\end{table*}

%% file: tables/moma_ft.tex
\begin{table*}[t]
\setlength{\tabcolsep}{0pt}
\setlength\extrarowheight{1pt}
\footnotesize
\begin{tabularx}{\linewidth}{l>{\centering\arraybackslash}X>{\centering\arraybackslash}XC{2mm}>{\centering\arraybackslash}X>{\centering\arraybackslash}XC{2mm}>{\centering\arraybackslash}X>{\centering\arraybackslash}XC{2mm}>{\centering\arraybackslash}X>{\centering\arraybackslash}X}
\toprule
\multirow{3}{*}{\vspace{0.1cm}\textbf{Model}} & \multicolumn{2}{c}{\textbf{Description}} & & \multicolumn{2}{c}{\textbf{Relationship}} && \multicolumn{2}{c}{\textbf{Action}} & & \multicolumn{2}{c}{\textbf{Total}} \\
\cmidrule{2-3} \cmidrule{5-6} \cmidrule{8-9} \cmidrule{11-12}
& \textbf{Accuracy} & \textbf{WUPS@0.9} & & \textbf{Accuracy} & \textbf{WUPS@0.9} & & \textbf{Accuracy} & \textbf{WUPS@0.9} & & \textbf{Accuracy} & \textbf{WUPS@0.9} \\
\midrule
InternVideo \cite{wang2022internvideo} & $42.15$ & $0.4215$ & & $36.77$ & $0.3980$ & & $71.12$ & $0.7112$ & & $45.91$ & $0.4804$ \\
mPLUG-2 \cite{mplug2} & $0.94$ & $0.0094$ & & $47.00$ & $0.7084$ & & $0.39$ & $0.0056$ & & $33.12$ & $0.4990$ \\
BLIP-2 \cite{blip2} & $62.90$ & $0.6290$ & & $77.34$ & $0.7864$ & & $72.55$ & $0.9286$ & & $75.73$ & $0.8278$ \\
Sevila \cite{yu2023self} & $63.60$ & $0.6360$ & & $78.92$ & $0.8218$ & & $74.60$ & $0.9453$ & & $77.19$ & $0.8442$ \\
\midrule
$\text{SGVLM}_{NoLoc}$ & $\textbf{67.01}$ & $\textbf{0.6701}$ & & $\underline{79.25}$ & $0.8216$ & & $74.71$ & $0.9560$ & & $\underline{77.60}$ & $0.8482$ \\
$\text{SGVLM}_{NoSG}$ & $65.33$ & $0.6533$ & & $78.73$ & $\underline{0.8263}$ & & $\underline{76.33}$ & $\underline{0.9763}$ & & $77.55$ & $\underline{0.8558}$ \\
SGVLM & $\underline{66.64}$ & $\underline{0.6664}$ & & $\textbf{81.36}$ & $\textbf{0.8435}$ & & $\textbf{77.06}$ & $\textbf{0.9771}$ & & $\textbf{79.66}$ & $\textbf{0.8688}$ \\
\bottomrule
\end{tabularx}
\caption{\textbf{Fine-tuned Performance Comparison of \ours~with Baselines on MOMA-QA Dataset.} $\text{SGVLM}_{NoLoc}$: An ablation of SGVLM where the frame localizer is removed and replaced with uniform frame sampling. $\text{SGVLM}_{NoSG}$: An ablation of SGVLM where the scene graph predictor is removed, and the model inferences solely on the frame embeddings.}
\label{tab:momaqa_ft}
\vspace{-1em}
\end{table*}

%% file: tables/nextqa.tex
\begin{table}[htbp!]
\small
\centering
\resizebox{\linewidth}{!}{%
    \begin{tabular}{@{}lccccc@{}}
        \toprule
        \textbf{Model} & \textbf{Causal} & \textbf{Temporal} & \textbf{Descriptive} & \textbf{Average} \\
        \midrule
        HGA \cite{hga} & $46.8$ & $52.1$ & $59.3$ & $50.4$ \\
        All-in-One \cite{wang2023all} & $48.0$ & $48.6$ & $63.2$ & $50.6$ \\
        Just Ask \cite{justask} & $49.6$ & $51.4$ & $63.1$ & $52.3$ \\
        MIST \cite{gao2023mist} & $54.6$ & $56.6$ & $66.9$ & $57.2$ \\
        HiTeA \cite{ye2023hitea} & $62.4$ & $58.3$ & $75.6$ & $63.1$ \\
        InternVideo \cite{wang2022internvideo} & $62.5$ & $58.5$ & $75.8$ & $63.2$ \\
        \midrule
        BLIP-2 \cite{blip2} & $72.9$ & $\underline{68.1}$ & $81.2$ & $72.6$ \\
        SeViLA \cite{yu2023self} & $\underline{74.2}$ & $\textbf{69.4}$ & $\underline{81.3}$ & $\underline{73.8}$ \\
        \midrule
        SGVLM & $\textbf{75.2}$ & $66.3$ & $\textbf{83.4}$ & $\textbf{74.3}$ \\
        \bottomrule
    \end{tabular}
    }
    \caption{\textbf{Comparison of \ours~with SoTA on NExT-QA.} We achieve comparable or slightly superior performance to existing methods on the NExT-QA dataset. This is noteworthy considering NExT-QA lacks explicit relationship and scene-graph oriented questions, underscoring the versatility of our approach.
    \label{tab:nextqa}
    }
\vspace{-2em}
\end{table}

%% file: sec/5_experiments.tex
\section{Experiments}

\input{tables/qvhl}

We evaluate our model against current state-of-the-art VideoQA models on \momaqa~and two public datasets: NExT-QA and QVHighlights. 

\subsection{Dataset \& Metrics}
\label{sec:metrics}

The \momaqa~dataset is evaluated on two metrics: \textbf{Accuracy} and \textbf{WUPS@0.9}. As \momaqa's questions are open ended, with test dataset $\mathbf{Q}$, the accuracy of the prediction $\hat{q}$ with respect to ground truth $q$ is given by:
    \begin{equation}
        acc = \frac{1}{|\mathbf{Q}|} \sum_{\mathbf{Q}} \frac{1}{|q|} \sum_{i=1}^{min(|\hat{q}|, |q|)} \mathbf{I}[\hat{q}_i = q_i].
    \end{equation}
WUPS is a soft measurement of accuracy used in multiple recent VideoQA datasets \cite{xiao2021next, yu2019activitynet}. The calculation method is detailed in Suppl.~\ref{sec:wups}.

The models are also evaluated on two public datasets: \textbf{NExT-QA} and \textbf{QVHighlights}. NExT-QA \cite{xiao2021next} is a VideoQA dataset focusing on causal and temporal action reasoning with 5,440 videos and 52,044 multiple-choice questions grouped into three categories: temporal, causal, and descriptive. We report categorical and overall accuracy. QVHighlights \cite{lei2021detecting} is a unified dataset for both moment retrieval and highlight detection. It contains 10,310 questions associated with 18,367 moments in 10,148 videos. We follow the evaluation metrics on the original paper and report R1, mAP on moment retrieval, and mAP and HIT@1 on highlight detection. 

\subsection{Experimental Setup}

We evaluate the performance of current state-of-the-art models on the datasets in both zero-shot and fine-tuned contexts. The \momaqa~results are compared against 4 popular models: InternVideo \cite{wang2022internvideo}, mPLUG-2 \cite{mplug2}, BLIP-2 \cite{blip2}, and SeViLa \cite{yu2023self}. The details of the experimental setup and training process are included in Suppl.~\ref{sec:exp_setup_appendix}.

%% file: tables/qvhl.tex
\begin{table}[t]
\footnotesize
\setlength{\tabcolsep}{0pt}
\begin{tabularx}{\linewidth}{@{\hspace{0mm}}p{2.2cm}p{0.85cm}<{\centering}p{0.85cm}<{\centering}p{1.0mm}<{\centering}p{0.85cm}<{\centering}p{0.85cm}<{\centering}p{0.85cm}<{\centering}p{1.0mm}<{\centering}p{0.85cm}<{\centering}p{0.8cm}<{\centering}}
\toprule
& \multicolumn{6}{c}{\textbf{Moment Retrieval}} & & \multicolumn{2}{c}{\textbf{HD}} \\
\cmidrule{2-7} \cmidrule{9-10}
& \multicolumn{2}{c}{R$1$} & & \multicolumn{3}{c}{mAP} & & \multicolumn{2}{c}{$\geq$ Very Good} \\
\cmidrule{2-3} \cmidrule{5-7} \cmidrule{9-10}
\vspace{-0.73cm}\textbf{Model} & @$0.5$ & @$0.7$ & & @$0.5$ & @$0.75$ & Avg. & & mAP & HIT@$1$ \\
\midrule
BeautyThumb \cite{song2016click} & $-$  & $-$  & & $-$  & $-$ & $-$  & & $14.36$ & $20.88$ \\
DVSE \cite{liu2015multi} & $-$ & $-$ & & $-$ & $-$ & $-$ & & $18.75$ & $21.79$ \\
MCN \cite{anne2017localizing} & $11.41$ & $2.72$ & & $24.94$ & $8.22$ & $10.67$ & & $-$ & $-$ \\
CAL \cite{escorcia2019temporal} & $25.49$ & $11.54$ & & $23.40$ & $7.65$ & $9.89$ & & $-$  & $-$  \\
CLIP \cite{radford2021learning} & $16.88$ & $5.19$ & & $18.11$ & $7.0$ & $7.67$ & & $31.30$  & $61.04$  \\
XML \cite{lei2020tvr} & $41.83$ & $30.35$ & & $44.63$ & $31.73$ & $32.14$ & & $34.49$ & $55.25$ \\
XML+ \cite{lei2021detecting} & $46.69$ & $33.46$ & & $47.89$ & $34.67$ & $34.90$ & & $35.38$ & $55.06$ \\
MDETR \cite{lei2021detecting} & $52.89$ & $33.02$ & & $54.82$ & $29.40$ & $30.73$ & & $35.69$ & $55.60$ \\
UniVTG \cite{lin2023univtg} & $58.86$ & $40.86$ & & $57.60$ & $35.59$ & $35.47$ & & $38.20$ & $60.96$ \\
UMT \cite{liu2022umt} & $56.23$ & $41.18$ & & $53.83$ & $37.01$ & $36.12$ & & $38.18$ & $59.99$ \\
QD-DETR \cite{moon2023query} & $\underline{62.40}$ & $\underline{44.98}$ & & $\textbf{62.52}$ & $\underline{39.88}$ & $\textbf{39.86}$ & & $\underline{38.94}$ & $\textbf{62.40}$ \\
SeViLA \cite{yu2023self} & $54.50$ & $36.50$ & & $-$ & $-$ & $32.30$ & & $-$ & $-$ \\
\midrule 
SGVLM & $\textbf{63.36}$ & $\textbf{46.30}$ & & $\underline{62.47}$ & $\textbf{42.00}$ & $\underline{39.82}$ & & $\textbf{39.17}$ & $\underline{62.26}$ \\
\bottomrule
\end{tabularx}
\captionsetup{font={small}}
\caption{\textbf{Moment Retrieval and Highlight Detection Results on QVHighlights Test Split.} We only include models not trained on additional video retrieval datasets (no extra training data). SGVLM (ours) and SeViLA are the only two VideoQA models.}
\vspace{-2em}
\label{tab:qvhl}
\end{table}

%% file: sec/6_discussion.tex
\section{Results \& Discussions}


In this section, we discuss the zero shot and fine-tuned VideoQA performance of various models on \momaqa~and NeXT-QA. We also report the results of the retriever alone on QVHighlights. Finally, we conduct a qualitative analysis of the results reported by \ours.

\begin{figure*}
    \centering
    \includegraphics[width=\linewidth]{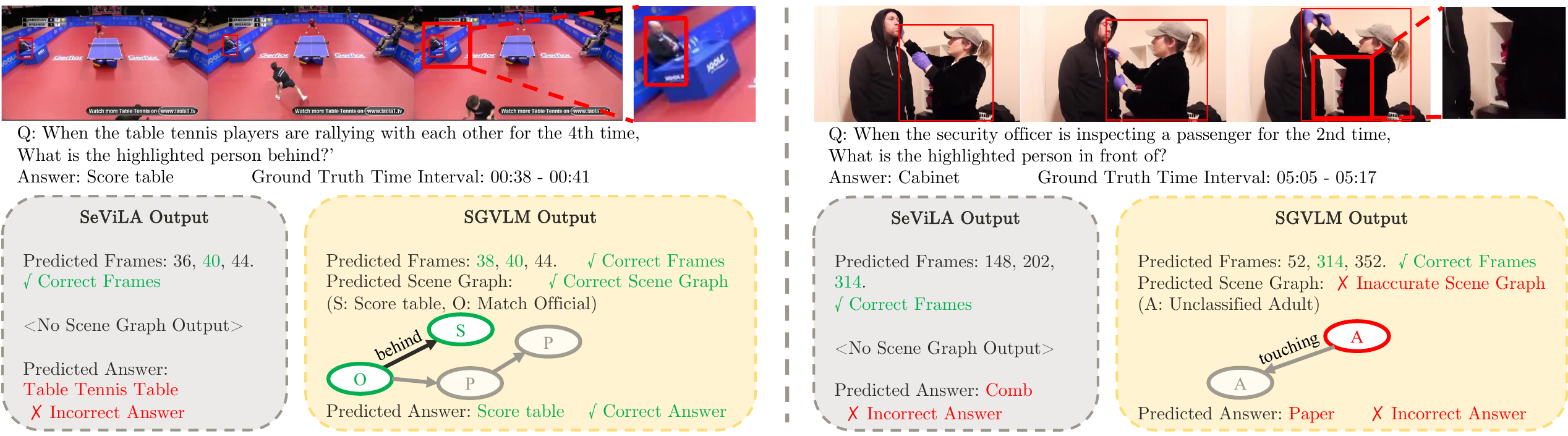}
    \caption{\textbf{Visualization Results of SGVLM with Previous SoTA (SeViLA) on \momaqa.} Left: An example where \ours~makes the correct prediction while SeViLA fails. Right: An example where both our model and SeViLA produce incorrect answers. We magnify the part from the frame that is relevant to the question for better readability.}
    \label{fig:error_analysis}
    \vspace{-1em}
\end{figure*}

\subsection{Zero Shot Results}

Table \ref{tab:momaqa_zs} shows the zero-shot metrics for the tested models. In our experiment, open vocabulary models surpass closed vocabulary ones in accuracy and WUPS, yet overall performances remain subpar. The best accuracy and WUPS@0.9 are 27.94\% (\ours) and 0.6023 (SeViLa) respectively. However, even open vocabulary models show limited zero-shot performance, with a maximum accuracy of 13.03\% and WUPS of 0.5045. These findings suggest a significant disparity between the \momaqa~dataset and the datasets on which these models were originally trained, highlighting that current VideoQA models struggle with the intricate relational dynamics featured in \momaqa, regardless of their model structures. The distinctive characteristics of \momaqa~therefore underscore its potential to introduce valuable diversity to the spectrum of VideoQA datasets available for advancing the field.

\subsection{Fine-tuned VideoQA Results}

\textbf{\momaqa.} Table~\ref{tab:momaqa_ft} presents a performance comparison of \ours~with several baselines on the MOMA-QA Dataset. \ours~outperforms the baseline methods across all metrics. Notably, in the Description and Relationship categories, SGVLM achieves accuracy scores of 66.64\% and 81.36\%, with corresponding WUPS@0.9 scores of 0.6664 and 0.8435, respectively. This represents a significant improvement of up to 3.04\% over the SeViLA model. Overall, SGVLM demonstrates the highest total accuracy at 79.66\% and a WUPS@0.9 score of 0.8688, suggesting robust video understanding performance across tasks. 

\noindent\textbf{NeXT-QA.} As shown in Table \ref{tab:nextqa}, SGVLM outperforms existing models on NeXT-QA. In the causal and descriptive questions, SGVLM sets new records with accuracies of 75.2\% and 83.4\%, respectively, exceeding the previous SoTA by up to 2.1\%. On average, SGVLM achieves an accuracy of 74.3\%, demonstrating its superior performance across different video understanding challenges.

\noindent\textbf{Ablations.} An ablation study of each component is shown in Table \ref{tab:momaqa_ft}, where we test how the model performs without the localizer and scene graph component. The ablations indicate that both parts contribute significantly to the model's performance. The SGVLM without the localizer ($\text{SGVLM}_{NoLoc}$) and without the scene graph predictor ($\text{SGVLM}_{NoSG}$) show reduced accuracy and WUPS@0.9 scores across all question categories (except \textit{description}) compared to the complete SGVLM model. Specifically, $\text{SGVLM}_{NoLoc}$ shows a slight decrease across most categories, while $\text{SGVLM}_{NoSG}$ shows a more pronounced decrease in the Relationship category, suggesting the scene graph predictor helps the model the most in the relationship category. These results underscore the importance of both frame localization and scene graph predictions in driving the model's superior performance.

\subsection{Frame Localization Results}

Our SGVLM model exhibits strong performance in Moment Retrieval and Highlight Detection tasks as shown in Table \ref{tab:qvhl}. SGVLM's capabilities are particularly evident in moment retrieval, where it tops the charts, exceeding the closest competitor by as much as 2.12\% in R1@0.5, R1@0.7, and mAP@0.7. In highlight detection, our model ranks second in mAP and leads in HIT@1, showcasing its precision in identifying video segments of interest. Notably, it outperforms the previously established state-of-the-art VideoQA model, SeViLA, by up to 9.8\%. These findings confirm SGVLM's effectiveness in accurately locating relevant video moments, highlighting its potential for real-world video analysis applications.

\subsection{Qualitative Analysis}

In our qualitative analysis, we compare SGVLM's output with the prior SoTA, SeViLA. Figure \ref{fig:error_analysis} (left) illustrates the task of identifying the object in front of the highlighted match official. SGVLM not only accurately selects relevant frames but also successfully constructs a scene graph, correctly recognizing the \textit{match official} as positioned \textit{behind} the \textit{score table}. In contrast, SeViLA, while identifying salient frames correctly, misinterprets the object as a \textit{table tennis table}. In this scenario, the use of scene graphs in SGVLM evidently contributes to its enhanced reasoning capabilities, which affirm the utility of structured semantic representations in complex VideoQA tasks.

In the right portion of Figure \ref{fig:error_analysis}, both models were assessed for their ability to correctly identify the object behind the highlighted person during a security inspection. Despite neither model successfully identifying the 'Cabinet' as the correct answer, SGVLM provides enhanced interpretability through its scene graph representation. The \textit{cabinet} is missing from SGVLM's scene graph, which suggests a limitation in the vision encoder's capability to recognize this occluded object. This interpretative result directs attention to potential enhancements in the vision encoding component of the model, indicating a clear pathway for future improvements in the model's overall ability.

%% file: sec/7_conclusion.tex
\section{Conclusion}

In this work, we introduce MOMA-QA, a VideoQA dataset that we hope will serve as a useful tool for advancing the fine-grained capabilities of VideoQA models by providing comprehensive frame-level annotations for spatio-temporally grounded QA. Towards this end, we introduce a novel video-language model, referred to as \ours. Our model uniquely leverages MOMA-QA's scene graph annotations for precise spatial relationship understanding and temporal localization annotations for effective frame selection. By integrating fine-grained video understanding with pre-trained large language models, we achieve a new state-of-the-art for VideoQA.

%% file: sec/X_suppl.tex
\clearpage
\setcounter{page}{1}
\maketitlesupplementary
\appendix

\section{Details of Scene Graph Predictor.}
\label{sec:sup_sg}

In this section, we detail the process of the scene graph predictor. Specifically, with object bounding boxes $B$, the object context $\mathbf{C}$ is first generated using a bidirectional LSTM layer:
\begin{equation}
    \mathbf{C} = \text{biLSTM}([\mathbf{f}_i; \mathbf{W}_{ctx}\mathbf{p}_i]_{i=1,\dots,n})
\end{equation}
where $\mathbf{W}_{ctx}$ is a learnable matrix. We then use a biLSTM layer and an MLP layer to encode each object into edge contexts:
\begin{align*}
    \mathbf{\hat{o}}_i &= \text{argmax}(\mathbf{W}_o \text{LSTM}([\mathbf{c}_i; \mathbf{\hat{o}}_{i-1}])) \\
    \mathbf{D} &= \text{MLP}(\text{biLSTM}([\mathbf{c}_i; \mathbf{W}_d \mathbf{\hat{o}}_{i-1}]))
\end{align*}
where $\mathbf{W}_d, \mathbf{W}_o$ are learnable matrices. In the end, for each pair of objects $(\mathbf{d}_i, \mathbf{d}_j)$, the scene graph feature $\mathbf{s}_{i, j}$ and the probability $\text{Pr}(x_{i \rightarrow j})$ is generated:
\begin{align*}
    \mathbf{s}_{i, j} = (\mathbf{W}_h \mathbf{d}_i)(\mathbf{W}_t \mathbf{d}_j) \\
    \text{Pr}(x_{i \rightarrow j}) = \text{softmax}(\mathbf{W}_r \mathbf{s}_{i, j})
\end{align*}

Finally, top k filtering is performed so that only the features with the top k probabilities $\mathbf{S} = \{\mathbf{s}_i, \dots, \mathbf{s}_k\}$ are saved for the next stage. The scene graph predictor is trained beforehand and kept frozen during the VideoQA training. 

\section{Calculation of WUPS@0.9}
\label{sec:wups}

With evaluation dataset $\mathbf{Q}$, the WUPS@0.9 score of the prediction $\hat{q}$ with respect to ground truth $q$ is given by
    \begin{align*}
        WUPS = \frac{1}{|Q|} \sum_{q \in Q} \min \{ \prod_{q_i \in A} \max_{\hat{q}_j} W_{\gamma}(q_i, \hat{q}_j), \\ 
        \prod_{\hat{q}_i} \max_{q_j} W_{\gamma}(\hat{q}_i, q_j) \}
    \end{align*}
    and $W_{\gamma}$ is given by
    \begin{align*}
        W_{\gamma}(q_i, \hat{q}_j) =
        \begin{cases} 
        W(q_i, \hat{q}_j) & \text{if } W(q_i, \hat{q}_j) \geq \gamma \\
        0.1W(q_i, \hat{q}_j) & \text{if } W(q_i, \hat{q}_j) < \gamma
        \end{cases}
    \end{align*}
    where we take $\gamma=0.9$ to calculate WUPS@0.9.

\section{Experimental Setup}
\label{sec:exp_setup_appendix}

We evaluate the models on \momaqa~in both fine-tuned and zero shot settings. The details of each experiment are included below.

\subsection{Zero Shot}
 
We evaluate the performance of current state-of-the-art models on \momaqa~in a zero-shot context. The experiment includes closed vocabulary models such as InternVideo \cite{wang2022internvideo} and mPLUG-2 \cite{mplug2}, as well as open vocabulary models like BLIP-2 \cite{blip2} and SeViLa \cite{yu2023self}. Each model is assessed on the test split of \momaqa~employing their respective optimal pre-trained parameters. For closed vocabulary models, answers are matched to the nearest word in the model's vocabulary when the precise answer falls outside its predefined vocabulary. 

For our model, \ours, the scene graph predictor is trained on the scene graph dataset Visual Genome \cite{xiao2020visual}; the frame localizer is trained on QVHighlights; and the full model is trained on NExT-QA \cite{xiao2021next} before being evaluated on \momaqa. 

\subsection{Fine-tuned}
 
We evaluate the same baseline models with our model on \momaqa~in a fine-tuned setting. We use the same initial weight as we used in the Zero Shot Experiment. Each model is trained on one computation node with four NVIDIA A6000s for a maximum of 5 epochs using the default hyperparameter settings from each model. The performance on the test dataset is reported. 

For our model, we use the same starting point as the zero shot experiment. The scene graph predictor and vision backbone are tuned on \momaqa. The frame localizer is also tuned while training the full model for VideoQA. 

\subsection{Experiments on Public Datasets}

In addition, we also evaluate models on NeXT-QA, a public dataset for video question answering, and QVHighlights, a public dataset for joint moment retrieval and highlight detections. NeXT-QA provides both multiple-choice and open-ended questions. For ease of comparison with baselines, we use the multiple-choice version of the dataset. Specifically, during evaluation, we take the probabilities for letters A, B, C, and D respectively, and choose the one with the highest probability as the prediction. This follows the standard practice employed in SeViLA \cite{yu2023self} and eliminates the probability that the model predicts unrelated tokens.

For QVHighlights, only the frame retriever component of \ours~is trained and evaluated. After tuning the model on the validation dataset, we train the model on a joint train-validation dataset and evaluate the model on the hidden test split. 

\section{Effect of Attention Masks}
\label{sec:att_masks}

\begin{table}[h]
\footnotesize
\setlength{\tabcolsep}{0pt}
\begin{tabularx}{\linewidth}{@{\hspace{0mm}}p{2.2cm}p{0.85cm}<{\centering}p{0.85cm}<{\centering}p{1.0mm}<{\centering}p{0.85cm}<{\centering}p{0.85cm}<{\centering}p{0.85cm}<{\centering}p{1.0mm}<{\centering}p{0.85cm}<{\centering}p{0.8cm}<{\centering}}
\toprule
& \multicolumn{6}{c}{\textbf{Moment Retrieval}} & & \multicolumn{2}{c}{\textbf{HD}} \\
\cmidrule{2-7} \cmidrule{9-10}
& \multicolumn{2}{c}{R$1$} & & \multicolumn{3}{c}{mAP} & & \multicolumn{2}{c}{$\geq$ Very Good} \\
\cmidrule{2-3} \cmidrule{5-7} \cmidrule{9-10}
\vspace{-0.73cm}\textbf{Model} & @$0.5$ & @$0.7$ & & @$0.5$ & @$0.75$ & Avg. & & mAP & HIT@$1$ \\
\midrule
Without Mask & $63.81$ & $47.35$ & & $62.21$ & $42.32$ & $40.60$ & & $39.69$ & $63.55$ \\
With Mask & $\textbf{64.65}$ & $\textbf{48.06}$ & & $\textbf{63.12}$ & $\textbf{43.19}$ & $\textbf{41.13}$ & & $\textbf{40.17}$ & $\textbf{64.19}$ \\
\bottomrule
\end{tabularx}
\captionsetup{font={small}}
\caption{\textbf{Ablation Results on QVHighlights Validation Split.} The best in each column is bolded.}
\label{tab:qvhl_ablation}
\end{table}

Table \ref{tab:qvhl_ablation} shows the effect of attention masks on the performance of the frame localizer. As shown in the table, the variant with the attention mask achieves a higher score on all metrics, with up to 0.91\% advantage on mAP@0.5. These results demonstrate the effectiveness of attention masks on the multi-modality input of the frame localizer. 